\documentclass[11pt]{article}

\usepackage[T1]{fontenc}
\usepackage[utf8]{inputenc}
\usepackage{mathpazo}
\usepackage{microtype}
\usepackage{geometry}
\usepackage{setspace}
\usepackage{amsmath,amssymb,amsthm,mathtools}
\usepackage{graphicx}
\usepackage{booktabs,tabularx,array}
\usepackage{float}
\usepackage[font=small,labelfont=bf]{caption}
\usepackage[round,authoryear]{natbib}
\usepackage[dvipsnames]{xcolor}
\usepackage{hyperref}
\usepackage[nameinlink,noabbrev]{cleveref}

\hypersetup{
  colorlinks=true,
  linkcolor=MidnightBlue,
  citecolor=MidnightBlue,
  urlcolor=MidnightBlue,
  pdftitle={Automation and the Depreciation of Adaptive Capacity},
  pdfauthor={Balaraju Battu}
}

\newtheorem{proposition}{Proposition}

\newcommand{\R}{\mathbb{R}}
\newcommand{\diff}{\mathop{}\!\mathrm{d}}
\newcommand{\proj}{\Pi_{[0,1]}}

\newcolumntype{Y}{>{\raggedright\arraybackslash}X}

\title{\textbf{Automation and the Depreciation of Adaptive Capacity}\\
\large Exploration, Rigidity, and Hysteresis}
\author{
Balaraju Battu$^{1,2}$\\
{\small $^{1}$European University Institute, Florence, Italy}\\
{\small $^{2}$New York University Abu Dhabi, UAE}
}
\date{September 2026}

\begin{document}

\maketitle

\begin{abstract}
Automation can raise current productivity while changing the human capabilities on which future adaptation depends. This paper models that second effect. Agents operate on a rugged task landscape and accumulate an endogenous stock of adaptive capacity through costly exploratory activity. When AI assistance substitutes for search, diagnosis, and experimentation, it reduces investment in that stock. With increasing returns to rebuilding adaptive capacity, the resulting dynamics exhibit a tipping point: a high-capacity equilibrium disappears once exploratory activity falls below a critical level. The economy then converges to a rigid state in which escape from locally efficient but obsolete routines becomes exponentially less likely. Because the low-capacity state remains locally stable after AI exposure is reversed, the model generates hysteresis and a distinct role for capability-rebuilding interventions. Decentralized adoption can exceed the social optimum when users capture current productivity gains but do not internalize the future value of exploratory capacity. The adverse result is not technological inevitability. AI systems that complement exploration reverse the core comparative static. The model therefore shifts attention from the quantity of AI use to the design of human--AI workflows and the adaptive capacities they reproduce.
\end{abstract}


\section{Introduction}

Artificial intelligence changes not only what workers produce but also how they arrive at an answer. In many knowledge-intensive tasks, generative and predictive systems shorten search, supply a plausible first solution, and remove intermediate steps that previously required diagnosis, experimentation, and revision. These tools can deliver substantial immediate productivity gains \citep{noy2023experimental,brynjolfsson2025generative,dell2026navigating}. Yet the omitted steps are not always waste. They can be investments in the ability to solve the next problem, especially when the next problem differs from those on which current tools and routines perform well.

This paper studies a simple question: what happens when automation substitutes for the activity through which adaptive capacity is reproduced? The conventional automation literature asks how technology changes tasks, employment, productivity, and the demand for skills \citep{acemoglu2018race,acemoglu2019automation,agrawal2023turing}. Here the stock of skill is itself affected by the mode of production. Independent search, error correction, and experimentation are costly inputs into current output, but they also maintain a capability that becomes valuable after a technological or organizational change. Automation can therefore create a wedge between current performance and future adaptability.

The mechanism has three parts. First, exploratory activity is a form of learning by doing \citep{arrow1962economic}. It maintains routines for framing unfamiliar problems, testing alternatives, and moving between domains. Evidence on cognitive offloading and AI-assisted work is consistent with the premise that external assistance can improve contemporaneous performance while weakening memory, effort, or independent skill formation \citep{risko2016cognitive,grinschgl2021consequences,macnamara2024ai,lee2025impact}. Second, the value of this capacity is greatest after a change in the environment. It may therefore remain hidden while familiar tasks are completed efficiently. Third, part of its value is social. Exploratory work produces public reasoning, mentorship, discarded alternatives, and cross-domain knowledge that private users need not internalize. Recent work on AI and collective knowledge emphasizes this externality from a complementary direction \citep{riochanona2024llm,lyu2025wikipedia,acemoglu2026ai}.

I formalize adaptive capacity as a state variable, \(z_t\in[0,1]\). It is regenerated by exploratory activity and depreciates when work becomes purely routinized. AI assistance, denoted \(a_t\), changes exploration according to
\[
s_t=\proj(s_0-\alpha a_t).
\]
The parameter \(\alpha\) is an effective substitution coefficient. When \(\alpha>0\), answer-first automation displaces human exploration. When \(\alpha<0\), AI complements exploration by making distant hypotheses, simulations, or domains easier to investigate. The sign of \(\alpha\), rather than AI use by itself, organizes the analysis.

The stock evolves according to
\[
\dot z_t=\eta s_t z_t^{\gamma}(1-z_t)-\rho(1-s_t)z_t,
\qquad \gamma>1.
\]
The first term captures capability formation through use; the second captures depreciation. The increasing-returns restriction \(\gamma>1\) is important. Rebuilding a weak capability often requires complementary routines, mentors, institutional memory, and a minimum base of prior competence. This creates an unstable threshold separating a high-capacity equilibrium from a low-capacity equilibrium. As substitutive AI lowers exploration, the high equilibrium and the threshold collide in a saddle-node bifurcation. Beyond that point, gradual depreciation becomes collapse.

Adaptive capacity matters because organizations and workers do not face a smooth, convex task space. They operate among locally coherent routines separated by switching costs, coordination problems, and knowledge barriers. I represent that environment by a rugged potential \(V(x)\). Local optimization pulls the system toward nearby routines, while exploratory mobility is proportional to \(z_t\). The escape rate from a local basin is approximately exponential in \(-1/z_t\). A modest decline in capacity can therefore have little visible effect on familiar work yet sharply reduce the ability to cross a barrier after the environment changes. This distinction between performance within a basin and mobility across basins is the central economic object of the paper.

The model yields five results. First, substitutive AI lowers the instantaneous accumulation of adaptive capacity at every interior state. Second, sufficiently strong substitution eliminates the high-capacity steady state. Third, falling capacity sharply raises metastable trapping on rugged task landscapes. Fourth, the low-capacity state can remain locally stable after AI intensity returns to its initial level. Reversal of exposure is then insufficient: recovery requires an intervention large enough to rebuild capacity above the unstable threshold. Fifth, decentralized adoption is excessive when users place less weight than society on the continuation value of adaptive capacity.

These results distinguish the argument from a generic claim that technology causes deskilling. The model permits both substitution and complementarity. An AI system that asks users to diagnose before revealing an answer, generates competing hypotheses, or lowers the cost of crossing disciplinary boundaries can increase \(s_t\). The same underlying capability can therefore produce different long-run effects under different workflows. The relevant design margin is whether AI compresses exploration into a private output or expands the user's feasible search set.

The paper contributes to three literatures. It adds an endogenous capability stock to models of automation and human--AI production. The closest recent work studies how AI can weaken incentives for learning, skill formation, or collective knowledge production \citep{caosun2026augmentation,aouad2026human,acemoglu2026ai}. The present model isolates a different state variable: mobility across unfamiliar task configurations. Information may remain abundant even as the capacity to recombine, test, and redirect it deteriorates. Second, the paper relates to strategic experimentation \citep{bolton1999strategic,keller2005strategic,keller2010strategic}. Instead of treating experimentation only as a way to learn an unknown payoff, I allow the ability to experiment to be produced by past experimentation. Third, the model connects economic path dependence \citep{arthur1988self,north1990institutions} to recent evidence on the cost of scientific and technological pivots \citep{hill2025pivot}. It shows how automation can endogenously change the effective height of those barriers by changing the explorer.

The rest of the paper proceeds as follows. \Cref{sec:model} presents the task landscape, exploration technology, and law of motion for adaptive capacity. \Cref{sec:dynamics} characterizes the tipping point, trapping, hysteresis, and adoption externality. \Cref{sec:empirics} develops empirical predictions and design implications. \Cref{sec:conclusion} concludes. Proofs and simulation details appear in the appendix.

\section{Model}
\label{sec:model}

\subsection{A rugged task landscape}

Let \(x_t\in\mathcal L\subseteq\R^n\) describe a worker's, firm's, or research team's current configuration: its problem representation, technical routine, organizational practice, or location in a knowledge space. The environment is summarized by a time-dependent potential \(V_t:\mathcal L\rightarrow\R\). Lower values of \(V_t\) are locally easier to sustain because they require less coordination, cognitive effort, or institutional switching. The potential is not welfare. A deep basin may represent a well-coordinated routine that is privately efficient under current conditions but inferior after the environment changes.

The state follows
\begin{equation}
\diff x_t=-\kappa\nabla V_t(x_t)\diff t+\sqrt{2Dz_t}\,\diff W_t,
\label{eq:landscape}
\end{equation}
where \(\kappa,D>0\) and \(W_t\) is an \(n\)-dimensional Brownian motion. The drift captures local adjustment toward nearby coherent routines. The diffusion term captures experimentation and variation. Adaptive capacity \(z_t\) scales exploratory mobility: when \(z_t\) is high, the system can sample alternatives and cross barriers; when \(z_t\) is low, it settles into a nearby basin and remains there.

This specification corrects an ambiguity that arises if responsiveness multiplies only the stabilizing drift. In \cref{eq:landscape}, lower adaptive capacity unambiguously reduces exploration. For a basin separated from an alternative by a barrier of height \(\Delta V\), the small-noise escape rate is approximately
\begin{equation}
r(z)\asymp K(z)\exp\left(-\frac{\kappa\Delta V}{Dz}\right),
\label{eq:kramers}
\end{equation}
where \(K(z)\) is a prefactor that varies more slowly than the exponential term \citep{gardiner2009stochastic}. Landscape ruggedness and adaptive capacity are therefore complements: the same loss of \(z\) is more consequential when barriers are high.

\subsection{AI assistance and exploratory activity}

Let \(a_t\in[0,1]\) denote the intensity of AI assistance and \(s_t\in[0,1]\) productive exploratory activity. Exploration includes independent diagnosis, generation of alternatives, testing, revision, and movement across unfamiliar domains. It is productive because it can improve current problem solving, but it is costly in time and effort.

AI changes exploration according to
\begin{equation}
s_t=s(a_t)=\proj(s_0-\alpha a_t),
\label{eq:exploration}
\end{equation}
where \(s_0\in(0,1]\) is baseline exploration. The projection keeps \(s_t\) in its feasible interval. The core analysis treats \(\alpha\) as fixed and focuses on the interior of the projection.

The coefficient \(\alpha\) is technological and institutional, not merely technical. An answer-first assistant that completes intermediate reasoning has \(\alpha>0\). A system that expands the set of hypotheses, requires the user to evaluate alternatives, or facilitates cross-domain search can have \(\alpha<0\). Later, I interpret
\begin{equation}
\alpha=\alpha(z,\mathcal I,\mathcal A),
\label{eq:endogenous-alpha}
\end{equation}
where \(\mathcal I\) denotes institutional routines and \(\mathcal A\) the architecture of interaction. Keeping \(\alpha\) fixed in the core model separates the stock dynamics from this design extension.

\subsection{The adaptive-capacity stock}

Adaptive capacity evolves as
\begin{equation}
\dot z_t=f(z_t;s_t)
=\eta s_t z_t^{\gamma}(1-z_t)-\rho(1-s_t)z_t,
\qquad \eta,\rho>0,\quad \gamma>1.
\label{eq:capacity}
\end{equation}
The regeneration rate \(\eta\) measures how effectively exploration builds future capacity. Saturation \((1-z_t)\) keeps the stock bounded. The depreciation rate \(\rho\) captures skill decay, routinization, loss of institutional memory, and dependence on external systems when exploration is absent.

The term \(z_t^{\gamma}\), with \(\gamma>1\), captures complementarity in capability formation. A system with established diagnostic routines, mentors, and domain knowledge converts an hour of exploration into more future capacity than a system in which those complements have disappeared. This assumption produces a genuine threshold and makes the claimed hysteresis mechanism explicit. If instead \(0<\gamma<1\), regeneration is relatively strongest near \(z=0\); that case produces smooth recovery rather than the collapse dynamics studied here.

The interval \([0,1]\) is forward invariant. At \(z=1\), \(\dot z=-\rho(1-s)\leq0\), and \(z=0\) is a steady state. Starting from a positive stock, the economy approaches zero asymptotically rather than reaching it in finite time.

\begin{table}[t]
\centering
\caption{Core objects}
\label{tab:notation}
\begin{tabularx}{\textwidth}{@{}>{$}l<{$}X@{}}
\toprule
\text{Object} & \text{Interpretation}\\
\midrule
x_t & Current task, knowledge, or organizational configuration\\
V_t(x) & Effective landscape of coordination and switching costs; not welfare\\
z_t\in[0,1] & Stock of adaptive capacity\\
s_t\in[0,1] & Current exploratory activity\\
a_t\in[0,1] & Intensity of AI assistance\\
\alpha & Effective substitution coefficient between AI and exploration\\
\eta,\rho & Regeneration and depreciation rates of adaptive capacity\\
\gamma>1 & Complementarity in capability formation\\
\Delta V & Barrier between locally stable configurations\\
\bottomrule
\end{tabularx}
\end{table}

\section{Dynamics and Economic Results}
\label{sec:dynamics}

\subsection{From gradual substitution to collapse}

The first result is local and does not require increasing returns.

\begin{proposition}[Immediate depreciation effect]
\label{prop:direct}
For an interior solution to \cref{eq:exploration}, an increase in AI assistance lowers the instantaneous growth of adaptive capacity whenever AI substitutes for exploration:
\[
\frac{\partial \dot z_t}{\partial a_t}
=-\alpha\left[\eta z_t^{\gamma}(1-z_t)+\rho z_t\right]<0
\]
for \(z_t\in(0,1)\) and \(\alpha>0\). If \(\alpha<0\), the inequality reverses.
\end{proposition}

The proposition separates current output from capability formation. A user can complete a task more quickly even while the growth rate of the capability stock falls. This gap is empirically important because current performance is a poor proxy for resilience to a future domain shift.

To characterize the global dynamics, hold exploration fixed at \(s\in(0,1)\). In addition to \(z=0\), any positive steady state satisfies
\begin{equation}
h(z)\equiv z^{\gamma-1}(1-z)
=q(s)\equiv\frac{\rho(1-s)}{\eta s}.
\label{eq:steadystate}
\end{equation}
The function \(h\) is single-peaked, with maximum at
\begin{equation}
z_m=\frac{\gamma-1}{\gamma},
\qquad
\bar h=\frac{(\gamma-1)^{\gamma-1}}{\gamma^{\gamma}}.
\label{eq:hmax}
\end{equation}

\begin{proposition}[Capacity threshold and collapse]
\label{prop:collapse}
Define
\begin{equation}
s_c=\frac{\rho}{\rho+\eta\bar h}.
\label{eq:scritical}
\end{equation}
If \(s>s_c\), the system has three steady states: the stable low-capacity state \(z=0\), an unstable threshold \(z_L(s)\), and a stable high-capacity state \(z_H(s)\), with
\[
0<z_L(s)<z_m<z_H(s)<1.
\]
At \(s=s_c\), the two positive steady states meet at \(z_m\). If \(s<s_c\), no positive steady state exists and every initial \(z\in(0,1]\) converges to zero.

For \(\alpha>0\), the corresponding AI threshold is
\begin{equation}
a_c=\frac{s_0-s_c}{\alpha},
\label{eq:a-critical}
\end{equation}
when this value lies in \([0,1]\).
\end{proposition}

Below \(a_c\), a sufficiently capable system can sustain the high steady state. The unstable branch is a minimum viable capability: falling below it sends the system toward the low state even though the high state still exists. As \(a\) rises, \(z_H\) falls and \(z_L\) rises. At \(a_c\), their basins collide and the high-capacity equilibrium disappears.

\begin{figure}[t]
\centering
\includegraphics[width=\textwidth]{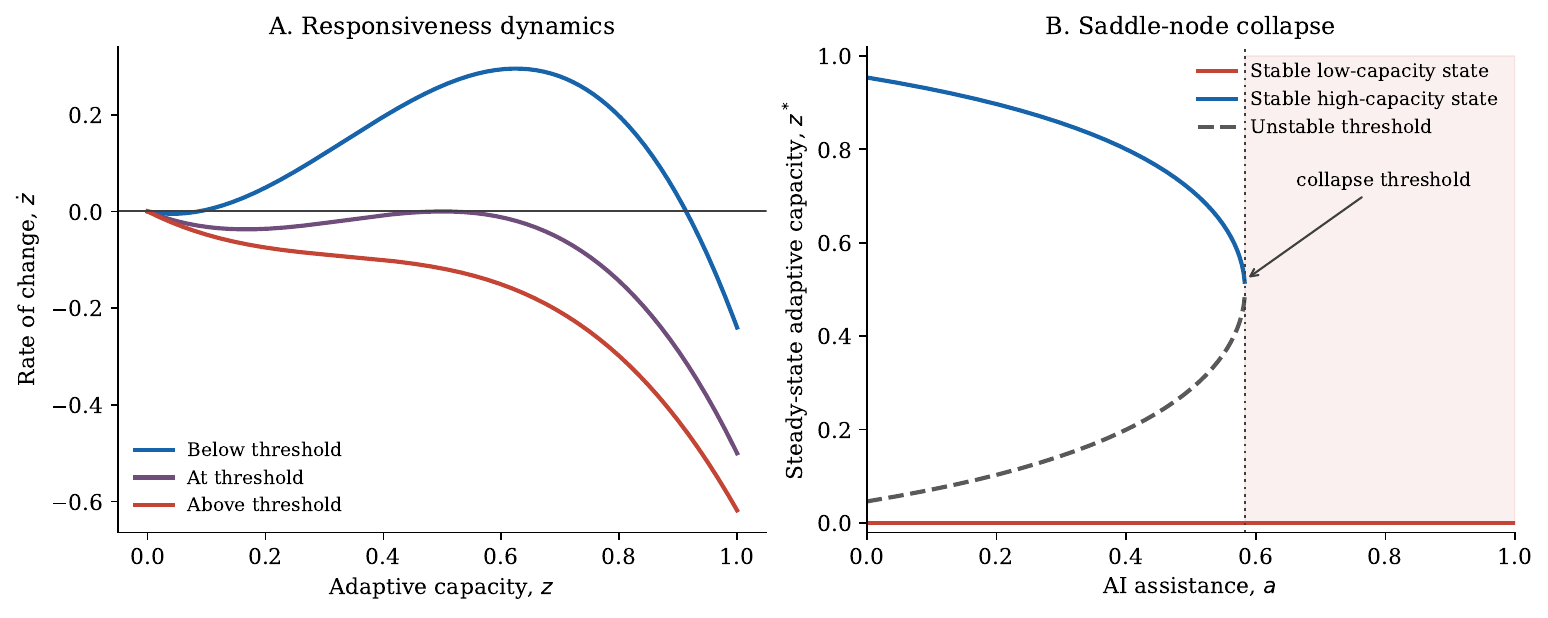}
\caption{\textbf{Adaptive-capacity dynamics and collapse.} Panel A plots \(\dot z\) below, at, and above the critical AI intensity. Panel B shows the steady states. Solid curves are stable; the dashed curve is the unstable threshold. Parameters are \(\eta=4\), \(\rho=1\), \(\gamma=2\), \(s_0=0.85\), and \(\alpha=0.60\), giving \(s_c=0.50\) and \(a_c\simeq0.583\).}
\label{fig:bifurcation}
\end{figure}

For the transparent benchmark \(\gamma=2\),
\[
z_{L,H}(s)=\frac{1\mp\sqrt{1-4q(s)}}{2},
\qquad
s_c=\frac{4\rho}{\eta+4\rho}.
\]
\Cref{fig:bifurcation} plots the vector field and these steady-state branches. Unlike a threshold imposed after the fact, the collapse point is generated by the capability law itself.

\subsection{Adaptive capacity and local trapping}

The stock \(z\) becomes economically relevant when the task landscape changes. Suppose the system occupies a basin that was previously attractive but is no longer globally desirable. Escape requires crossing a barrier \(\Delta V\).

\begin{proposition}[Metastable trapping]
\label{prop:trapping}
In the small-noise regime of \cref{eq:landscape}, the escape rate from a basin is increasing in adaptive capacity and decreasing in the barrier height. In particular, the leading exponential term satisfies
\[
\frac{\partial\log r(z)}{\partial z}
\simeq\frac{\kappa\Delta V}{Dz^2}>0.
\]
Hence the effect of a decline in \(z\) is amplified on more rugged landscapes.
\end{proposition}

The result distinguishes local efficiency from global mobility. A low-\(z\) organization can optimize effectively inside an established workflow; the problem is its inability to reach another workflow when the relative value of basins changes. Coordination costs, installed infrastructure, specialization, and legitimacy constraints all contribute to \(\Delta V\). This interpretation connects increasing returns and institutional lock-in \citep{arthur1988self,north1990institutions} with measured penalties for moving across distant knowledge domains \citep{hill2025pivot}.

\begin{figure}[t]
\centering
\includegraphics[width=\textwidth]{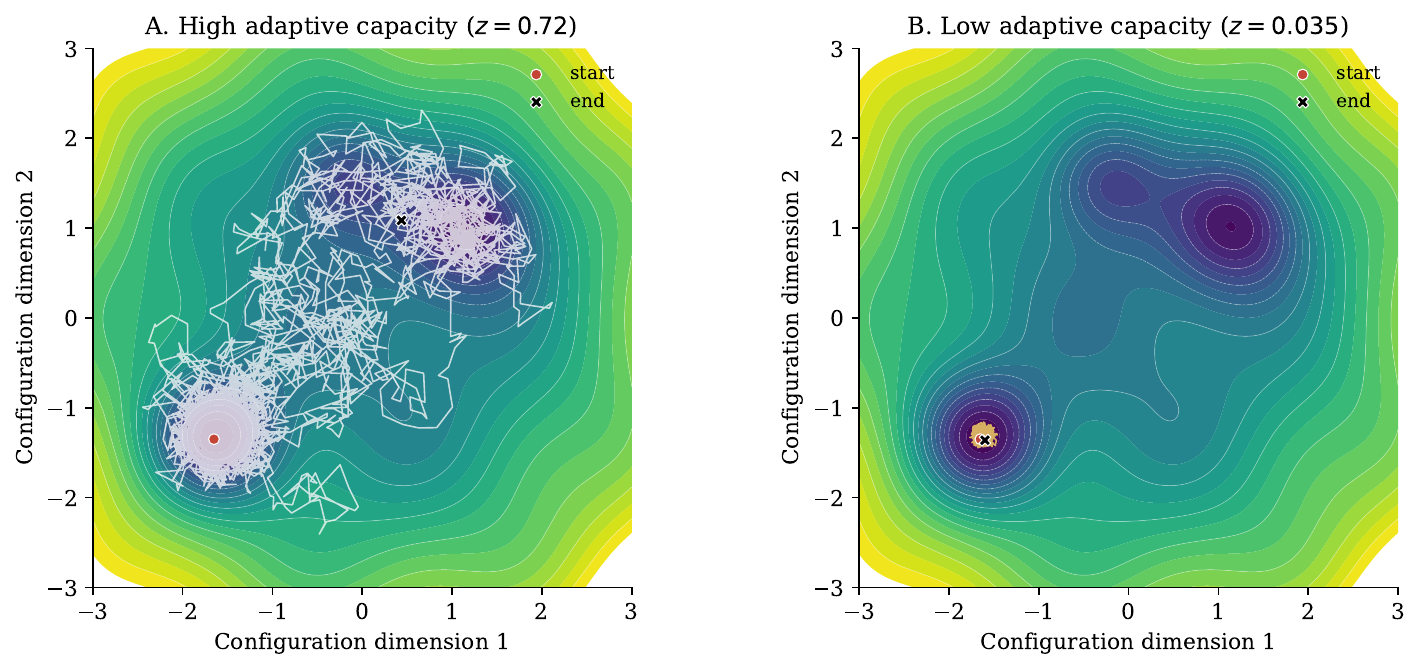}
\caption{\textbf{Adaptive capacity on a rugged task landscape.} Both panels use the same potential, initial condition, and shock sequence. Higher \(z\) scales exploratory mobility in \cref{eq:landscape}, permitting traversal between basins. Low \(z\) produces persistent localization. The figure is illustrative rather than calibrated.}
\label{fig:landscape}
\end{figure}

\Cref{fig:landscape} illustrates the distinction on the same rugged potential. At high adaptive capacity, stochastic exploration reaches several basins. At low capacity, the trajectory remains concentrated near its initial attractor. The landscape is identical across panels; only the explorer changes.

\subsection{Hysteresis and recovery}

The coexistence of a stable low state and a stable high state makes exposure history matter. Begin with \(s_0>s_c\) and \(z_0>z_L(s_0)\). A temporary increase in substitutive AI lowers \(s\). If it persists long enough to push \(z\) below the unstable threshold associated with the restored baseline, simply reducing AI use does not reverse the decline.

\begin{proposition}[Hysteresis]
\label{prop:hysteresis}
Suppose \(s_0>s_c\), \(\alpha>0\), and an interval of high AI exposure leaves adaptive capacity at \(\tilde z<z_L(s_0)\). After assistance returns to its initial level, \(a=0\), the stock converges to zero rather than to \(z_H(s_0)\). Recovery requires either a direct capability investment that raises \(z\) above \(z_L(s_0)\) or a temporary change in \(s\), \(\eta\), or \(\rho\) that moves the separatrix below the current stock.
\end{proposition}

This is structural hysteresis, not merely slow adjustment. The initial environment again admits a high-capacity equilibrium, but the system is now in the basin of the low equilibrium. Policies that only remove the original exposure can therefore fail. Apprenticeship, protected practice, human-first diagnosis, and deliberate retraining have a different role: they rebuild the state variable or change its law of motion.

\begin{figure}[t]
\centering
\includegraphics[width=\textwidth]{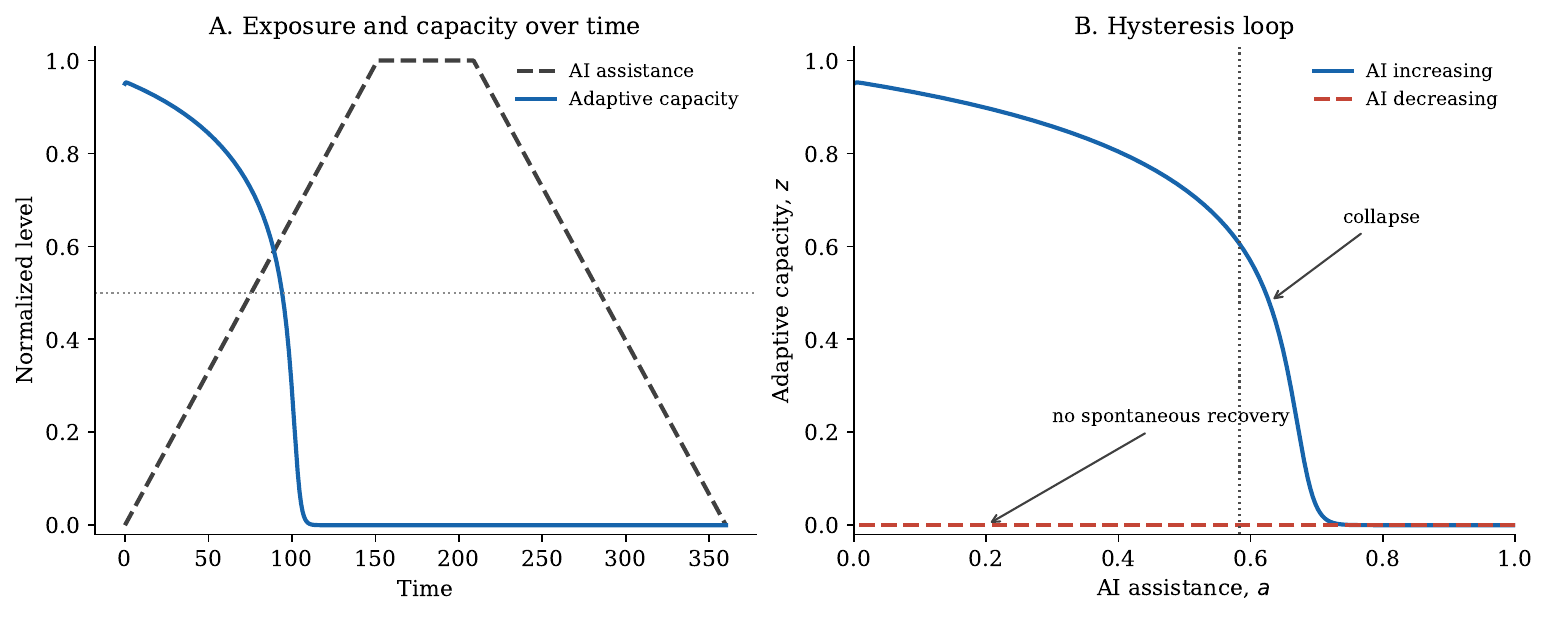}
\caption{\textbf{Hysteresis under temporary AI exposure.} AI assistance rises, remains high, and returns to zero. The high-capacity state disappears after the saddle-node threshold is crossed. Once \(z\) enters the basin of the low state, reversing AI intensity does not generate spontaneous recovery. Parameters match \cref{fig:bifurcation}.}
\label{fig:hysteresis}
\end{figure}

\subsection{The adoption externality}

Let \(B(a)\) denote the net current benefit of AI assistance. On the high-capacity branch, write the private and social objectives as
\begin{align}
J^P(a)&=B(a)+\beta_P H(z_H(a)),\\
J^S(a)&=B(a)+\beta_S H(z_H(a)),
\label{eq:welfare}
\end{align}
where \(H'(z)>0\) and \(0\leq\beta_P<\beta_S\). The term \(H(z)\) is a reduced-form continuation value: future performance after domain shifts, the maintenance of public reasoning, and spillovers from a population that can diagnose unfamiliar problems. Private decision-makers may value some of this capacity, but less than society.

\begin{proposition}[Excess adoption under substitution]
\label{prop:welfare}
Suppose \(\alpha>0\), both objectives in \cref{eq:welfare} are strictly concave with unique interior optima on the high-capacity branch, and \(z_H'(a)<0\). Then the social optimum satisfies
\[
a^S<a^P.
\]
If the private optimum crosses the collapse threshold, accounting for the discontinuous loss of adaptive capacity can widen the wedge.
\end{proposition}

At the private optimum, the marginal current benefit is balanced against the privately valued loss of capacity. The planner attaches a larger weight to the same loss. The wedge is not an assertion that all exploration should be retained. Some exploratory effort is redundant, and moderate AI use can raise both objectives. The result is that a decentralized system can remove more exploration than is socially desirable.

\begin{figure}[t]
\centering
\includegraphics[width=0.82\textwidth]{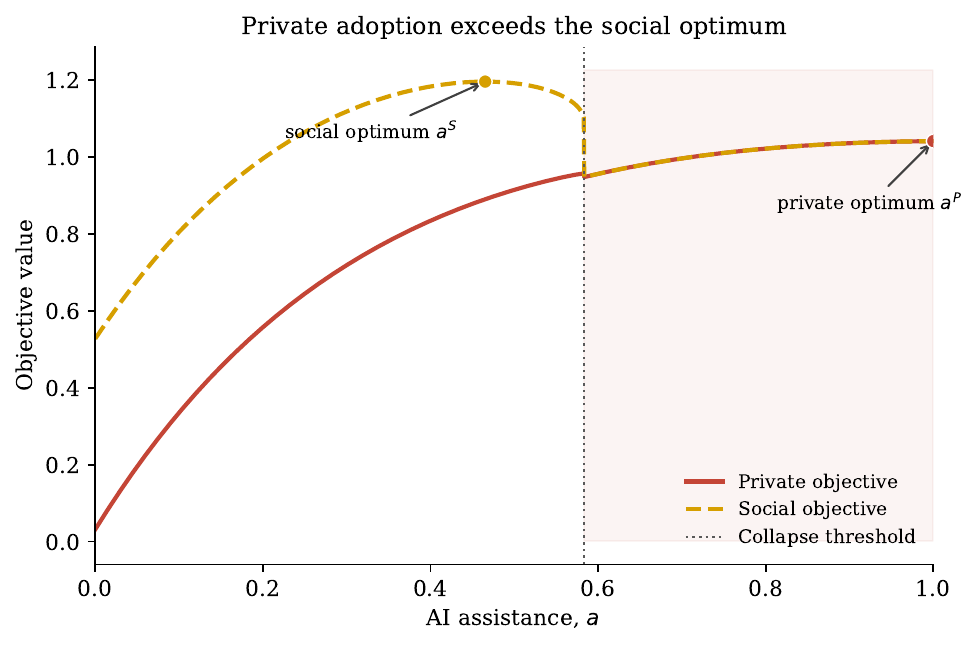}
\caption{\textbf{Private and social AI adoption.} The private objective places little weight on the continuation value of adaptive capacity. The social objective peaks earlier and falls discretely when the high-capacity branch disappears. Functional forms and parameters are reported in Appendix~\ref{app:numerics}.}
\label{fig:welfare}
\end{figure}

\subsection{When AI complements exploration}

The preceding collapse results condition on \(\alpha>0\). For \(\alpha<0\), Proposition~\ref{prop:direct} reverses: AI raises exploration, shifts \(q(s)\) downward, lowers the unstable threshold, and raises the high-capacity steady state. This case describes systems that generate contrasting hypotheses, expose uncertainty, support simulation, or translate across domains without replacing evaluation by the user.

Equation \cref{eq:endogenous-alpha} makes the design margin explicit. Three features can change the effective coefficient. First, user capacity matters: the same output may complement an expert's search while substituting for a novice's still-developing routines. Second, institutions determine whether human diagnosis and public reasoning remain part of the workflow. Third, interface architecture determines whether AI supplies a terminal answer or expands the set of alternatives the user must assess. These channels can also create feedback. If low \(z\) makes answer substitution more attractive, then \(\partial\alpha/\partial z<0\), strengthening collapse. If low capacity triggers protected training or more Socratic assistance, institutional feedback can work in the opposite direction.

The normative implication is therefore conditional. The model does not recommend less capable AI. It recommends evaluating AI systems by the joint production of current output and future human adaptability.

\section{Empirical Implications and Design}
\label{sec:empirics}

The model separates three objects that are often conflated: current exploratory activity \(s_t\), the latent stock of adaptive capacity \(z_t\), and observed output on familiar tasks. A credible empirical design should measure all three or introduce a domain shift that reveals \(z_t\).

\subsection{Predictions}

The model generates five testable predictions.

\begin{enumerate}
\item \textit{Performance and capacity can diverge.} Substitutive AI raises output on familiar tasks while reducing unaided performance on novel tasks after sufficient exposure.

\item \textit{Effects accumulate with exposure.} Two workers with the same current AI access can respond differently to a novel problem because their prior exposure histories imply different \(z_t\).

\item \textit{Ruggedness magnifies depreciation.} The effect of a decline in capacity is larger when adaptation requires a distant technological, scientific, or organizational pivot rather than a nearby adjustment.

\item \textit{Withdrawal is asymmetric.} Removing AI assistance need not restore unaided performance at the rate at which it deteriorated. Recovery should be especially weak after exposure crosses the estimated basin threshold.

\item \textit{Architecture changes the sign.} Answer-first systems and exploration-oriented systems can produce similar current output but different effects on branching, hypothesis diversity, and subsequent unaided adaptation.
\end{enumerate}

\begin{table}[t]
\centering
\caption{From model objects to empirical measures}
\label{tab:measurement}
\small
\begin{tabularx}{\textwidth}{@{}>{\raggedright\arraybackslash}p{0.16\textwidth}>{\raggedright\arraybackslash}p{0.38\textwidth}Y@{}}
\toprule
Model object & Candidate measures & Diagnostic variation\\
\midrule
Exploration \(s_t\) & Number and diversity of attempted solutions; discarded code branches; revision entropy; competing hypotheses; cross-domain search & Randomize answer-first versus question-first, adversarial, or simulation-oriented assistance\\[3pt]
Adaptive capacity \(z_t\) & Unaided performance after a domain shift; time to diagnose an unfamiliar failure; successful pivots; recovery speed & Remove assistance after different exposure durations, then introduce novel tasks\\[3pt]
Ruggedness \(\Delta V\) & Technological distance; retraining or switching costs; coordination intensity; disciplinary distance & Interact treatment with ex ante task distance or an unanticipated regime change\\[3pt]
Substitution \(\alpha\) & Change in human exploratory actions per unit of AI use & Compare workflows holding model capability and nominal access fixed\\[3pt]
Hysteresis & Gap between capacity on increasing and decreasing exposure paths & Ramp exposure up and down; add a randomized capability-rebuilding treatment\\
\bottomrule
\end{tabularx}
\end{table}

\subsection{Identification}

One design is a longitudinal experiment in which teams receive the same AI model through different interfaces. An answer-first treatment delivers a recommended solution immediately. An exploration-oriented treatment first elicits a diagnosis, generates several alternatives, and requires evaluation before revealing a recommendation. Current task quality identifies the productivity effect. Subsequent unaided performance on structurally novel tasks identifies changes in \(z\). Varying exposure duration can reveal nonlinearity; withdrawing access and randomizing retraining can distinguish ordinary persistence from basin-dependent hysteresis.

A field design can combine staggered AI adoption with an external domain shift. Pre-shift production measures estimate current gains. Post-shift adaptation speed, exploratory breadth, and the probability of a successful pivot reveal capacity. The model predicts that cumulative substitutive exposure matters even after conditioning on current use. Scientific work offers observable measures of distance and search: topic transitions, citation diversity, abandoned projects, and cross-field recombination. Evidence that AI expands impact while narrowing the distribution of topics would be consistent with a productivity--mobility distinction \citep{hao2026artificial}, although it would not by itself identify the stock mechanism.

The strongest test is a path experiment. Randomly assign units to the same final level of AI assistance through different histories. If final outcomes depend only on current assistance, the stock model is unnecessary. If novel-task performance differs by history and a targeted rebuilding treatment closes the gap, the joint evidence supports adaptive-capacity depreciation and hysteresis.

\subsection{Institutional design}

The model points to interventions that act on different margins. Limiting AI intensity changes \(a\). Redesigning the interface changes \(\alpha\). Protected practice changes \(s\). Apprenticeship, mentoring, and retraining raise \(\eta\) or directly increase \(z\). Modular standards and lower switching costs reduce \(\Delta V\). These policies are not equivalent. After collapse, reducing \(a\) may be ineffective unless paired with a measure that crosses the capability threshold.

This distinction helps avoid a false choice between unrestricted automation and blanket prohibition. Institutions can preserve human-first diagnosis for high-consequence edge cases, require alternative generation before convergence, maintain public reasoning and documentation, and periodically test unaided response to unfamiliar tasks. The purpose is not to preserve effort for its own sake. It is to retain the option value of a system that can still move when its current optimum ceases to be one.

\section{Conclusion}
\label{sec:conclusion}

Automation can improve what a system does today while depreciating its ability to do something different tomorrow. The paper formalizes that tension by treating adaptive capacity as an endogenous stock produced through exploration. When AI substitutes for the search and diagnostic activity that renews this stock, the effect is initially gradual. With complementary inputs to capability formation, however, gradual substitution can cross a saddle-node threshold, eliminate the high-capacity equilibrium, and generate persistent rigidity.

Embedding the stock in a rugged task landscape gives the mechanism economic force. Declining capacity sharply reduces the probability of escaping a locally coherent routine, especially when adaptation requires a distant pivot. The same local productivity statistics can therefore conceal very different levels of systemic resilience. Because the low-capacity state can remain stable after exposure is reversed, prevention and recovery are different policy problems.

The analysis is deliberately parsimonious. It abstracts from heterogeneous workers, strategic firms, endogenous task creation, and the political economy of automation. Those extensions matter, but they do not alter the basic accounting identity: exploratory activity can produce both a current answer and a future ability. An evaluation that records only the first output will overstate the gains from substituting it away.

The central design question is not whether AI is used. It is whether human--AI systems reproduce the capacity to explore. Answer-first automation can compress that capacity; exploration-oriented systems can expand it. Long-run evaluation should therefore measure performance after novelty, not only productivity within the familiar basin.

\section*{Acknowledgements and disclosures}

The author received no financial support for this research, authorship, or publication. The author designed the research, developed the model, and wrote the manuscript.

\clearpage
\bibliographystyle{plainnat}
\bibliography{references}

\clearpage
\appendix

\section{Proofs}
\label{app:proofs}

\subsection*{Proof of Proposition~\ref{prop:direct}}

On the interior of \cref{eq:exploration}, \(\partial s/\partial a=-\alpha\). Differentiating \cref{eq:capacity} first with respect to \(s\) gives
\[
\frac{\partial f}{\partial s}
=\eta z^{\gamma}(1-z)+\rho z>0
\]
for \(z\in(0,1)\). Applying the chain rule yields the expression in the proposition. \hfill\(\square\)

\subsection*{Proof of Proposition~\ref{prop:collapse}}

Factor \cref{eq:capacity} as
\[
f(z;s)=z\left[\eta s z^{\gamma-1}(1-z)-\rho(1-s)\right].
\]
For \(s<1\) and \(\gamma>1\), \(f_z(0;s)=-\rho(1-s)<0\), so \(z=0\) is locally stable. Positive equilibria solve \(h(z)=q(s)\), as defined in \cref{eq:steadystate}. Differentiation gives
\[
h'(z)=z^{\gamma-2}\left[(\gamma-1)-\gamma z\right].
\]
Thus \(h\) rises on \((0,z_m)\), falls on \((z_m,1)\), and attains \(\bar h\) at \(z_m\). There are two, one, or zero positive roots according as \(q(s)\) is below, equal to, or above \(\bar h\). At a positive root,
\[
f_z(z;s)=\eta s z h'(z),
\]
so the lower root is unstable and the upper root stable. Solving \(q(s_c)=\bar h\) gives \cref{eq:scritical}; combining it with \cref{eq:exploration} gives \cref{eq:a-critical}. \hfill\(\square\)

\subsection*{Proof of Proposition~\ref{prop:trapping}}

Equation \cref{eq:landscape} is an overdamped diffusion with effective noise intensity \(Dz\). In the small-noise limit, the leading Kramers term for passage over a barrier \(\Delta V\) is \cref{eq:kramers}. Differentiating its logarithm while treating the prefactor as lower order gives
\[
\frac{\partial\log r(z)}{\partial z}
\simeq \frac{\kappa\Delta V}{Dz^2}>0,
\qquad
\frac{\partial\log r(z)}{\partial \Delta V}
\simeq-\frac{\kappa}{Dz}<0.
\]
\hfill\(\square\)

\subsection*{Proof of Proposition~\ref{prop:hysteresis}}

When the baseline \(s_0>s_c\) is restored, Proposition~\ref{prop:collapse} implies \(f(z;s_0)<0\) on \((0,z_L(s_0))\). Hence any restored state \(\tilde z\) in that interval moves monotonically toward the stable equilibrium at zero. The high equilibrium remains in a different basin. Raising \(z\) above \(z_L(s_0)\), or changing parameters so that the unstable threshold falls below \(\tilde z\), moves the system into the basin of \(z_H\). \hfill\(\square\)

\subsection*{Proof of Proposition~\ref{prop:welfare}}

On the high branch, implicit differentiation of \(h(z_H)=q(s(a))\) gives \(z_H'(a)<0\) for \(\alpha>0\): \(q\) rises with \(a\), while \(h'\) is negative at \(z_H\). At the private optimum \(a^P\),
\[
B'(a^P)+\beta_P H'(z_H(a^P))z_H'(a^P)=0.
\]
Evaluating the derivative of social welfare at \(a^P\) yields
\[
(J^S)'(a^P)
=(\beta_S-\beta_P)H'(z_H(a^P))z_H'(a^P)<0.
\]
Strict concavity therefore implies \(a^S<a^P\). \hfill\(\square\)

\section{Numerical Illustration}
\label{app:numerics}
\enlargethispage{9\baselineskip}

All figures are generated by the included script \texttt{generate\_figures.py}. They illustrate the analytical mechanisms and are not calibrated estimates.

\subsection{Bifurcation and hysteresis}

The benchmark uses
\[
\eta=4,\qquad \rho=1,\qquad \gamma=2,\qquad s_0=0.85,\qquad \alpha=0.60.
\]
Thus \(s_c=0.50\), \(a_c=0.583\overline{3}\), and \(z_m=0.50\). For the hysteresis experiment, AI assistance rises linearly from zero to one over the first 42 percent of the horizon, remains at one until 58 percent, and then declines linearly to zero. The capacity equation is integrated from \(z_0=0.95\).

\subsection{Rugged landscape}

The illustrative potential is
\begin{align*}
V(x,y)={}&0.09(x^2+y^2)
-1.25\exp\left[-\frac{(x+1.65)^2+(y+1.35)^2}{0.46}\right]\\
&-1.05\exp\left[-\frac{(x-1.25)^2+(y-1.05)^2}{0.62}\right]
-0.72\exp\left[-\frac{(x+0.25)^2+(y-1.55)^2}{0.42}\right]\\
&+0.11\sin(2.5x)\cos(2.1y).
\end{align*}
The Euler--Maruyama simulations set \(\kappa=1\), \(D=0.34\), and \(\Delta t=0.006\). Both panels use the same initial condition and random-number seed. Adaptive capacity is \(z=0.72\) in the high-capacity panel and \(z=0.035\) in the low-capacity panel.

\subsection{Adoption illustration}

The net current benefit is
\[
B(a)=1.18(1-e^{-3a})-0.08a^2.
\]
The continuation value is \(H(z)=z^2\), with \(\beta_P=0.035\) and \(\beta_S=0.58\). Before the collapse point the economy follows \(z_H(a)\); after the high branch disappears, the illustration assigns the low state \(z=0\). These choices produce an interior social optimum below the threshold and a private optimum at higher AI intensity. The ordering, rather than the numerical locations, is the object of interest.

\end{document}